\documentclass[10pt,twocolumn,letterpaper]{article}

\usepackage{cvpr}              
\definecolor{cvprblue}{rgb}{0.21,0.49,0.74}
\usepackage[pagebackref,breaklinks,colorlinks,allcolors=cvprblue]{hyperref}

\def\paperID{} 
\def\confName{CVPR}
\def\confYear{2026}

\title{PaLMR: Towards Faithful Visual Reasoning via Multimodal Process Alignment}

\makeatletter
\def\blfootnote{\gdef\@thefnmark{}\@footnotetext}
\makeatother

\author{
    Yantao Li$^{1,2,3}$,  Chenyang Yan$^1$, Qiang Hui$^{2,3}$, Fang Zhao$^{2,3,\dagger}$, Kanzhi Cheng$^1$,  \\
    Chao Tan$^{2,3}$, Huanlin Gao$^{2,3}$, Jianbing Zhang$^{1,*}$, Kai Wang$^{2,3}$, Xinyu Dai$^1$, Shiguo Lian$^{2,3,*}$\vspace{0.4em}\\
    $^1$ National Key Laboratory for Novel Software Technology, Nanjing University\\
    $^2$ Data Science \& Artificial Intelligence Research Institute, China Unicom\\
    $^3$ Unicom Data Intelligence, China Unicom \\
    {\tt\small li\_yantao@smail.nju.edu.cn}\qquad
    {\tt\small \{zjb,daixinyu\}@nju.edu.cn}\qquad
    {\tt\small \{huiq,zhaof50,liansg\}@chinaunicom.cn}
    \vspace{-1em}
}
\usepackage[ruled]{algorithm2e}
\usepackage{xcolor}

\begin{document}

\twocolumn[{
\renewcommand\twocolumn[1][]{#1}
\maketitle
\begin{center}
    \centering
    \vspace{5pt}
    \captionsetup{type=figure}
    \includegraphics[width=1.0\textwidth]{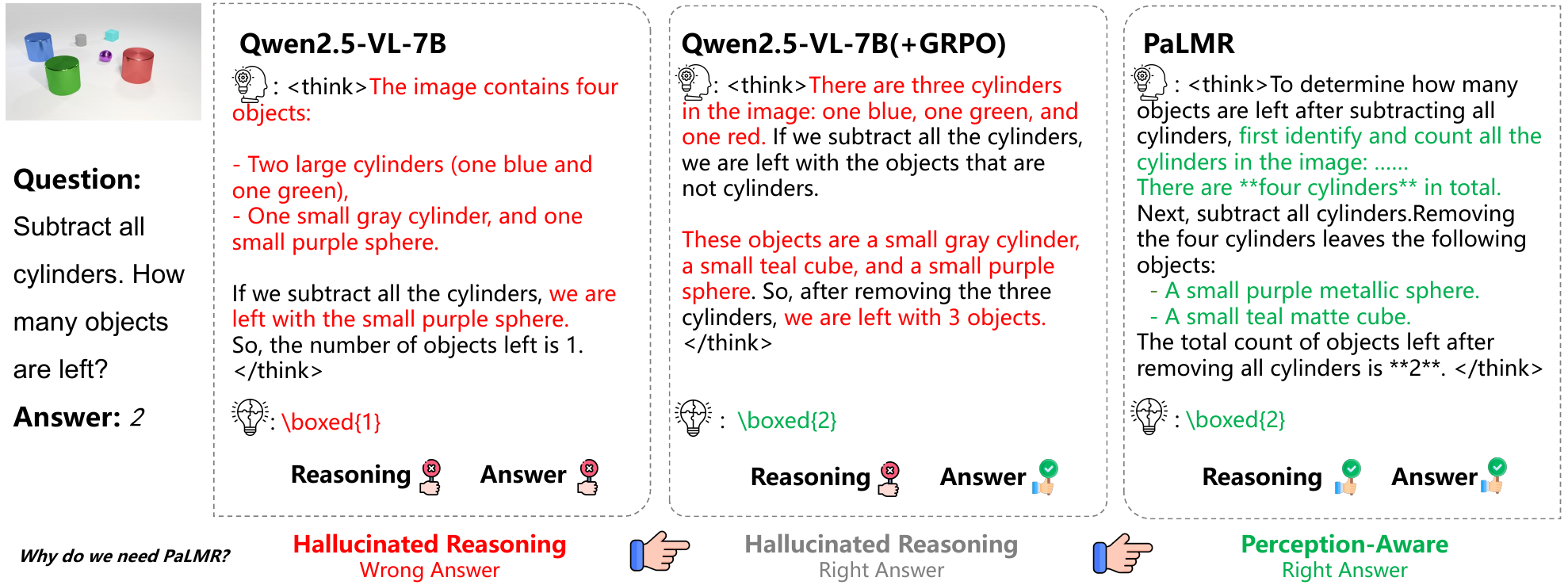}
    \captionof{figure}{\textbf{Comparison of reasoning behaviors among baseline models, baseline models(+GRPO) and PaLMR on a visual reasoning sample.} As shown, PaLMR demonstrates perception-aware reasoning and produces faithful answers by process-level perception alignment, addressing the hallucinated reasoning issue in prior models.}
    \label{fig:teaser}
\end{center}
}]

\begin{abstract}

\blfootnote{$^*$ Corresponding authors. $^\dagger$ Project leader. }


Reinforcement learning has recently improved the reasoning ability of Large Language Models (LLMs) and Multimodal LLMs (MLLMs), yet prevailing reward designs emphasise final-answer correctness and consequently tolerate process hallucinations—cases where models reach the right answer while misperceiving visual evidence. We address this process-level misalignment with \textbf{PaLMR} (Process Alignment for Multimodal Reasoning), a framework that aligns not only outcomes but also the reasoning process itself. \textbf{PaLMR} comprises two complementary components: a perception-aligned data layer that constructs process-aware reasoning data with structured pseudo-ground-truths and verifiable visual facts, and a process-aligned optimisation layer that constructs a hierarchical reward fusion scheme with a process-aware scoring function to encourage visually faithful chains-of-thought and improve training stability. Experiment on Qwen2.5-VL-7B shows that our approach substantially reduces reasoning hallucinations and improves visual reasoning fidelity, achieving state-of-the-art results on HallusionBench while maintaining strong performance on MMMU, MathVista, and MathVerse. These findings indicate that PaLMR offers a principled and practical route to process-aligned multimodal reasoning, advancing the reliability and interpretability of MLLMs.
\end{abstract}
    
\section{Introduction}

Multimodal large language models(MLLMs) have recently achieved remarkable progress in visual reasoning and perception-language understanding. Building upon reinforcement learning(RL) successes in large language models(LLMs) such as DeepSeek-R1~\cite{guo2025deepseekr1} and R1-Zero~\cite{zhou2025r1zero}, researchers have extended outcome-based optimization to multimodal settings, outcome many promissing models, like MM-Eureka~\cite{meng2025mm}, OpenVLThinker~\cite{deng2025openvlthinker} and Perception-R1~\cite{yu2025perception}, which improve answer accuracy across benchmarks such as MMMU~\cite{yue2024mmmu} and MathVista~\cite{lu2023mathvista}. However, most existing reward mechanisms focus solely on what the model answers, while overlooking how it reasons. This often leads to reasoning hallucinations — models obtain correct answers through visually inconsistent reasoning. For example, it may claim ``three cups are on the table'' on the chain-of-thought(CoT)~\cite{wei2022chain} when four are clearly visible, yet still predict the right answer based on textual priors.  

Such hallucinations reveal a core limitation of current multimodal reinforcement learning: rewards are primarily \emph{outcome-oriented}, offering supervision only for final correctness while neglecting the \emph{faithfulness} of process reasoning steps. 
This observation highlights the need for \textit{process-level alignment}: ensuring that the reasoning trajectory remains consistent with visual evidence at every step. Previous reinforcement learning with visual reasoning methods~\cite{meng2025mm, deng2025openvlthinker} mainly optimize textual reasoning results, while visual reward models such as VisualPRM~\cite{wang2025visualprm} and VRPRM~\cite{chen2025vrprm} rely on human preference comparisons between visually correct and incorrect samples.

In contrast, our method links perceptual accuracy to process-level reasoning via a verifiable binary visual-textual consistency metric. We propose \textbf{PaLMR} (\textbf{P}rocess \textbf{aL}ignment for \textbf{M}ultimodal \textbf{R}easoning) to enforce visual faithfulness throughout the reasoning process, moving beyond outcome-only correctness. PaLMR comprises two components: a perception-aligned data layer (\textbf{PaDLayer}) that generates visually grounded reasoning samples, and a process-aligned optimization layer (\textbf{PaOLayer}) that reinforces trajectory coherence. To stabilize policy learning, \textbf{PaOLayer} employs a hierarchical reward fusion scheme integrating perception and outcome scores. We implement this via \textbf{Vision-Guided Group Relative Policy Optimization (V-GRPO)}, embedding visual-consistency rewards into the \emph{RL} objective. V-GRPO prioritizes process-level perception feedback, effectively unifying perception fidelity and reasoning quality.

Extensive experiments on multiple benchmarks demonstrate that \textbf{PaLMR} consistently improves reasoning faithfulness and visual consistency while maintaining competitive answer accuracy. Notably, our method significantly reduces hallucination rates in reasoning and achieves state-of-the-art visual alignment performance among multimodal reasoning models. 

Our main contributions are summarized as follows:
\begin{itemize}
    \item We introduce the \textbf{PaLMR} framework, a faithful multimodal process alignment framework that enforces reasoning-process faithfulness by unifying perception-aligned data construction layer and process-aligned optimization layer.
    \item We propose \textbf{V-GRPO} training paradigm that enhances visual faithfulness by incorporating a perception-aware scoring strategy and integrating visual cues into the \textbf{GRPO} framework, forming a hierarchical reward mechanism that jointly optimizes reasoning accuracy and perception consistency.
    \item  Experiments show that \textbf{PaLMR} significantly outperforms baseline models in both alignment faithfulness and reasoning quality.
\end{itemize}

\section{Related Work}

\subsection{Multimodal Chain-of-Thought Reasoning}
The Chain-of-Thought (CoT)~\cite{wei2022chain} paradigm has become central to reasoning enhancement in both textual and multimodal large language models~\cite{wu2023multimodallargelanguagemodels}. Early works such as \textit{Visual Thoughts}~\cite{cheng2025visualthoughts} and \textit{CoT-VLA}~\cite{zhao2025cotvla} demonstrated that decomposing multimodal reasoning into explicit intermediate steps significantly improves interpretability and performance on benchmarks like ScienceQA~\cite{lu2022learn} and A-OKVQA~\cite{schwenk2022okvqa}. \textit{Kam-CoT}~\cite{kamcot} and \textit{LLaVA-CoT}~\cite{llava-cot} extended this paradigm to vision--language models by introducing explicit visual reasoning chains. Subsequent works such as \textit{Self-Consistency}~\cite{wang2022selfconsistency} and \textit{Tree-of-Thought}~\cite{yao2023tree} encouraged multi-path exploration and reflection. While test-time compute methods~\cite{snell2024scaling,cheng-etal-2025-vision,sun2025mm} incorporated self-reflection signals with CoT to improve reasoning performance. Together, these studies established CoT reasoning as a fundamental approach for stepwise multimodal understanding.

\subsection{Reinforcement Learning for Multimodal Reasoning}
Reinforcement-learning frameworks have been widely used to improve reasoning robustness and factuality.
\textit{DeepSeek-R1}~\cite{guo2025deepseekr1} and \textit{R1-Zero}~\cite{zhou2025r1zero} introduced group-relative policy optimization to incentivize structured reasoning behavior in large language models. Extending this idea to vision--language models, \textit{Vision-R1-Zero}~\cite{huang2025visionr1zero} and \textit{R1-VL}~\cite{zhang2025r1vl} applied rule-based and visual-feedback-guided optimization to enhance multimodal reasoning consistency. These methods established a strong foundation for post-training alignment of reasoning behavior but primarily relied on outcome-level feedback: judging only the correctness of the final answer without explicitly supervising intermediate reasoning or perceptual faithfulness.

\begin{figure*}[htbp]
    \centering
    \includegraphics[width=1.0\linewidth]{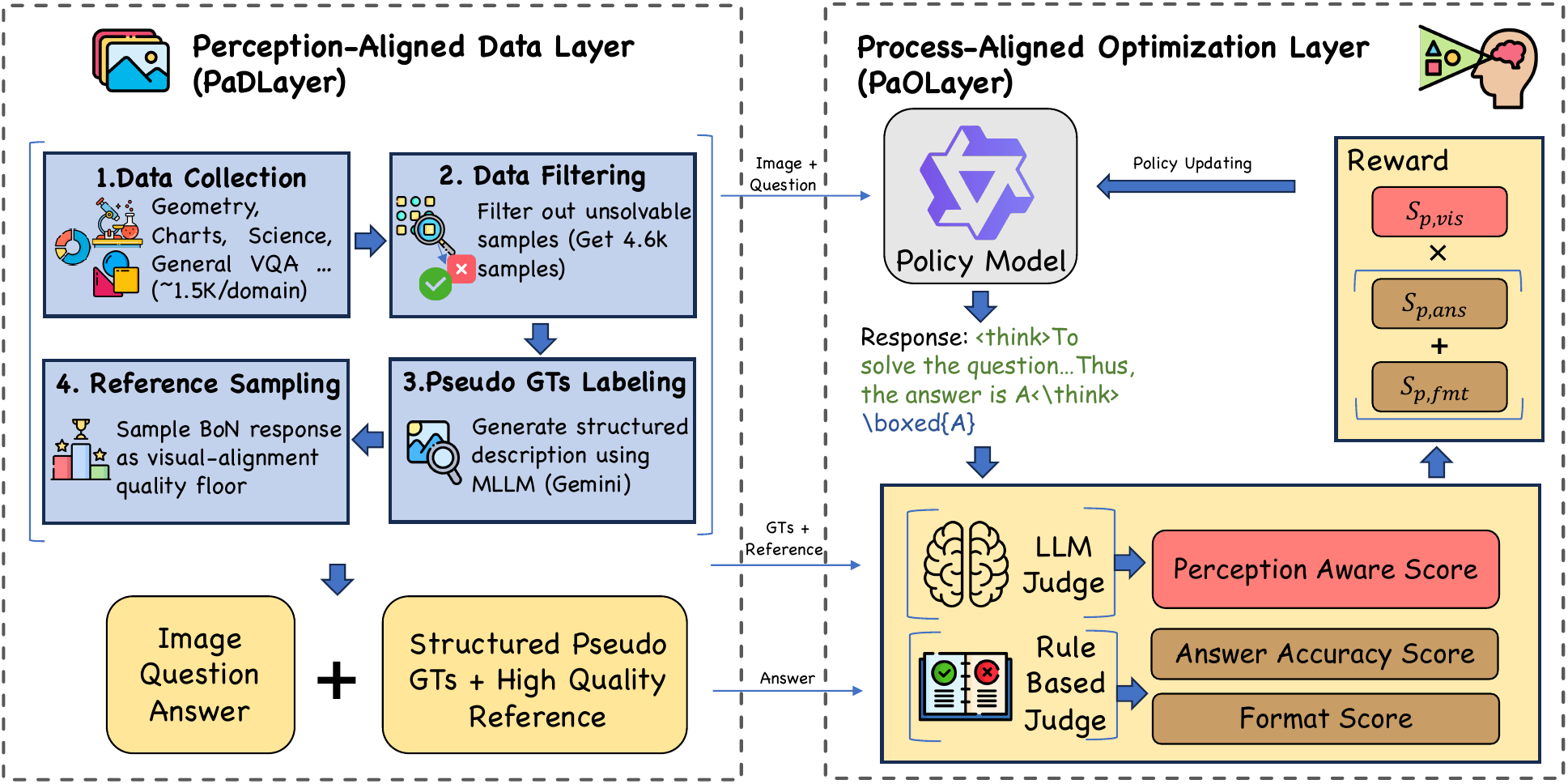}
    \caption{\textbf{Overview of the proposed PaLMR framework.} The model adopts a two-layer architecture: (a) the \textbf{Perception-Aligned Data Layer (PaDLayer)} builds process-aware multimodal data with structured pseudo ground truths and verifiable visual facts; and (b) the \textbf{Process-Aligned Optimization Layer (PaOLayer)} integrates perception-aware, answer, and format rewards into GRPO to enforce visually faithful and logically coherent reasoning.}
    \label{fig:palmr_overview}
\end{figure*}

\subsection{Process Reward Models for Multimodal Reasoning}
Recent work has shifted from outcome-based RL to process-level supervision, rewarding models for faithful intermediate reasoning steps. \textit{VRPRM}~\cite{chen2025vrprm} pioneered visual reasoning process reward modeling with a two-stage training strategy, achieving over 100\% relative improvement in Best-of-N(BoN) selection. 
\textit{ViLPRM} and its benchmark \textit{ViLBench}~\cite{tu2025vilbench} evaluated visual--language process rewards at scale, while \textit{VisualPRM}~\cite{wang2025visualprm} provided fine-grained step-level evaluation using a BoN selector. Further variants such as \textit{Mm-PRM}~\cite{du2025mmprmenhancingmultimodalmathematical}, \textit{PRM-BAS}~\cite{hu2025prmbasenhancingmultimodalreasoning}, and \textit{GM-PRM}~\cite{zhang2025gmprm} supervise each reasoning step to reduce error accumulation; \textit{GM-PRM} even treats the PRM as an active collaborator that can refine reasoning chains via the Refined-BoN strategy. 
Other models, including \textit{DreamPRM}~\cite{cao2025dreamprm}, \textit{ER-PRM}~\cite{zhang2024entropy} and \textit{EDU-PRM}~\cite{cao2025process}, enhance calibration through entropy regularization or ranking-based supervision.Weakly and self-supervised approaches such as \textit{FreePRM}~\cite{sun2025freeprm} explore pseudo-label and self-reward signals to reduce annotation costs. Unified frameworks—\textit{Unified Multimodal Chain-of-Thought Reward Model}~\cite{wang2025unifiedcot} and \textit{Unified Reward Model} for Multimodal Understanding and Generation~\cite{zhang2025unifiedrm}—generalize process reward modeling across reasoning, understanding, and generation tasks. These studies highlight a growing shift toward process-aware reward modeling, in which models are evaluated and optimized based on the quality, coherence, and perceptual grounding of their intermediate reasoning steps.

\section{Method}
\label{sec:method}
We propose \textbf{PaLMR}, a unified framework that enhances visual faithfulness by aligning perception and reasoning via process-level scoring and reinforcement optimization. As in Fig.~\ref{fig:palmr_overview}, PaLMR has two interdependent layers: a \textbf{PaDLayer} that constructs verifiable multimodal data, and a \textbf{PaOLayer} that enforces process alignment through hierarchical scoring and vision-guided GRPO.

\subsection{Preliminaries}
A visual reasoning task often provides a textual question along with an image and asks an MLLM to provide a response sequence for the question. Formally, given a query sequence $\mathbf{x} = [x_1, x_2, \dots, x_n]$ and corresponding image $\mathbf{I}$, the MLLM model $\mathcal{M}$ generates sequence $\mathbf{y} = [y_1, y_2, \dots, y_m]$, where $x_i$ and $y_i$ are individual tokens. The whole responce $\mathbf{y}$ is sampled from conditional distribution $p_\theta(\cdot|\mathbf{x, I})$, as $p_\theta(\mathbf{y}|\mathbf{x, I}) = \prod_{j=1}^m p_{\theta}(y_j|\mathbf{x, I}, y_1, y_2, \dots, y_{j-1})$.



\textbf{Group Relative Policy Optimization(GRPO)\cite{shao2024deepseekmath}:} GRPO is a widely used policy gradient RL algorithm that leverages intra-group relative performance to optimize the policy model.
During Training, for each query $x$, the model samples a group of $G$ responses, and GRPO computes the relative advantage of the group based on its rewards. $\{r_1, r_2, \dots, r_G\}$ as follows:
$$
    A^i = \frac{r_i - \text{mean}(\{r_i\}_{i=1}^G)}{\text{std}(\{r_i\}_{i=1}^G)}
$$
Using the advantage values, GRPO optimizes the model with PPO-clip loss in terms of:
\begin{equation}
J_{\text{GRPO}}(\theta) = -\mathbb{E}_{\substack{x\sim \mathcal{D} \\ y\sim \pi_\theta}} \left[\text{Avg}({ \min\left( \psi_i A_i,\ \text{clip}(\psi_i, 1\pm\epsilon) A_i \right) } \right]
\label{eq:grpo_objective}
\end{equation}
where $\psi_i$ and $A_i$ donate per-token policy ratio and advantage on response $y_i$, and $\text{Avg}(\cdot)$ means the average over group and response length.
By adjusting the components of the reward function, practitioners can steer the policy toward behaviors that yield higher rewards.
In RLVR (reinforcement learning from verifiable rewards), it is common to use rule-based, verifiable signals such as exact-answer correctness to provide reliable rewards and mitigate the risk of reward hacking.

However, when exact-answer correctness is used as the sole reward signal during optimization of Multimodal Large Language Models (MLLMs), the resulting optimization tends to prioritize textual accuracy at the expense of visual perception integrity.
This text-centric reward fails to adequately capture the importance of visual information in multimodal reasoning tasks.
To address this limitation, we propose a pairwise, visually aware score that explicitly aligns the model's responses with visual fidelity during reinforcement learning.

\subsection{Perception-Aligned Data Layer (PaDLayer)}\label{sec:PaDlayer}
\textbf{PaDLayer} establishes a verifiable basis for process-level alignment through a four-step procedure (see Figure \ref{fig:palmr_overview}). We begin by uniformly sampling 1,500 instances from FineVision~\cite{wiedmann2025finevisionopendataneed} across multiple reasoning domains, including geometry, charts, science, and OCR, to mitigate distributional bias.

We then apply a data filtering strategy based on learnability to select informative examples. Specifically, we perform stochastic rollouts under the \textbf{PaOLayer} policy and remove outliers that are consistently incorrect or unstable, as well as trivial cases with excessively high accuracy that do not contribute meaningful reinforcement learning signals. We further exclude formatting-incompatible problems using rule-based matching. This process results in approximately 4,700 retained instances.

To establish verifiable targets, we use \textit{Gemini} to generate structured pseudo ground-truths for each sample. These captions enumerate objects, spatial relations, and visual attributes, thereby converting images into symbolic representations. We then sample a semantically coherent reasoning trajectory from the policy model using a Best-of-N selection, which serves as the reference-response baseline for subsequent visual-faith optimization.

\subsection{Process-Aligned Optimization Layer (PaOLayer)}\label{sec:PaOLayer}

After the training sets are ready, we use them to enhance visual process reasoning by coupling perception-aware scoring with \textbf{GRPO}, forming a \textbf{Vision-Guided GRPO (V-GRPO)} training strategy. \textbf{V-GRPO} are constructed by assessing the visual faithfulness of reasoning trajectories through pairwise comparison and producing reliable binary visual fidelity scores. This integration ensures that models learn to reason accurately, perceiving not only the final correctness.

\textbf{Perception-Aware Scoring:} A common approach to quantifying visual consistency is first to extract visual claims $\mathbf{Z}$ from the model's chain of thought and then compute a consistency score by comparing them against a set of ground-truth facts. This score, denoted as the visual-aligned score, can be integrated into the reward function to penalize hallucinated or missing visual details.
However, our preliminary experiments revealed a significant limitation of this point-wise evaluation paradigm: it is susceptible to the intrinsic biases of the "LLM-as-judge". We observed that this method tends to yield high accuracy only for reasoning paths that are already essentially correct, but struggles to provide a reliable judgment when imperfect or partially correct trajectories appear. To mitigate evaluator bias while retaining the flexibility of LLM-as-judge, we replace point-wise scoring with pairwise comparisons, using the LLM to judge which trajectory demonstrates more faithful and coherent reasoning. To ensure stable, consistent comparisons throughout the RL training process, we first sample a target trajectory $\tau_{\text{target}}$ during training. With a given high-quality reference trajectory $\tau_{\text{ref}}$ sourcing from \emph{PaDLayer}, LLM determines whether the former is superior to the latter. This finally yields a binary visual fidelity score: 
\begin{equation} 
S_{p,vis}(\tau) = 
\mathbb{I}(\tau_{\text{target}} \succ \tau_{\text{ref}}) \label{eq:pairwise_reward} \end{equation}
 where $\mathbb{I}(\cdot)$ is the indicator function and the relation $\succ$ is a preference induced by the LLM-as-judge given the ground-truth facts. $S_{v,vis}(\tau)$ equals $1$ if the target is preferred and $0$ otherwise.
\begin{figure}[tp]
    \centering
    \includegraphics[width=1.0\linewidth]{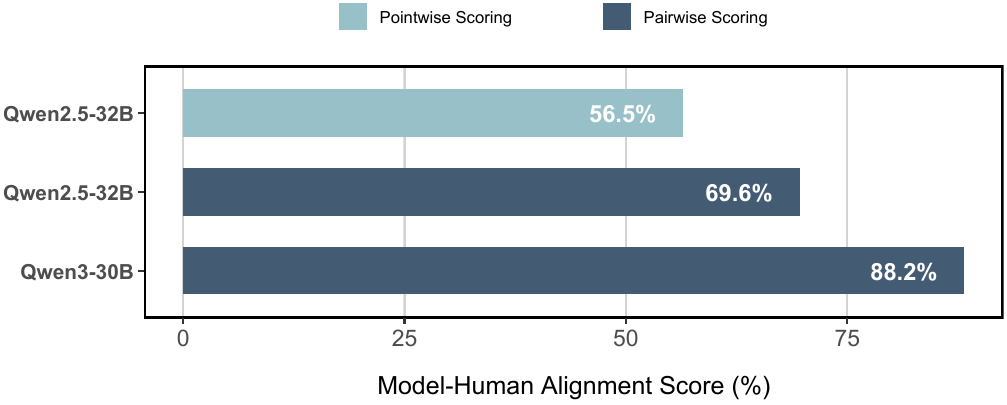}
    \caption{\textbf{Model-Human alignment ratio in identifying visual perception errors.} The evaluation assesses visual perception errors in 100 randomly sampled responses generated from the Geo3K dataset with Qwen2.5-VL-7B, and using Qwen2.5-32B, Qwen3-30B as judgement model.
    }
    \label{fig:model-human-judge}
\end{figure}
 
Figure \ref{fig:model-human-judge} illustrates that the pairwise evaluation paradigm achieves a significantly higher human alignment ratio compared to the point-wise method. Furthermore, utilizing powerful models, such as Qwen3\cite{yang2025qwen3}, can lead to over 88\% model-human alignment. Such a high level of concordance between LLM-as-judge and human evaluators provides a robust and reliable binary signal for optimizing the model's perceptual grounding.
 
This approach effectively combines the scalability of LLM-as-judge with the enhanced accuracy of comparative judgment, providing a robust, human-aligned signal for perceptual fidelity. However, it remains an isolated evaluation metric unless tightly coupled with the optimization process. So we embed $S_p^{\text{pair}}$ directly into the reward structure of GRPO, forming a Vision-Guided GRPO (V-GRPO). 

\textbf{V-GRPO:} To enforce visual alignments as a prerequisite for successful task completion, we design a hierarchical reward function for our \textbf{V-GRPO}. It prioritizes visual cues, following the principle of Perception-Aware Scoring, e.g., $S_p$. A trajectory deemed visually unfaithful ($S_p =0$) receives no reward for its final answer, regardless of its accuracy. So it strictly penalizes any reasoning that contains visual hallucinations or misinterpretations. The total reward $R_{V-GRPO}(\tau)$ is formulated as a hierarchical combination of scores: 
\begin{equation} 
R_{V-GRPO}(\tau) =  S_{p,\text{vis}}(\tau) \cdot (\alpha S_{p, \text{ans}}(\tau) + (1-\alpha) S_{p,\text{fmt}}(\tau))\label{eq:hierarchical_reward_concise} 
\end{equation} 
where $S_{p,\text{vis}}(\tau)$ is the binary visual fidelity score, $S_{p, \text{ans}}(\tau)$ and $S_{p, \text{fmt}}(\tau)$ are rule-based scores for answer-accuracy and format-correctness and $\alpha$ is a balancing coefficient. Specifically, $S_{p,\text{vis}}(\tau)$ enforces visual fidelity and holds the highest priority—if a trajectory contains perceptual errors, the entire reward is set to zero. $S_{p,\text{ans}}(\tau)$ encourages task-level correctness, and $S_{p,\text{fmt}}(\tau)$ ensures that the model’s outputs remain structurally and syntactically consistent. By integrating this reward mechanism into the GRPO algorithm, we propose V-GRPO. This approach explicitly forces the model first to learn to "see correctly" before learning to "reason correctly," thereby promoting the development of more faithful and reliable models.


  

\begin{table*}[htbp]
    \centering
    \begin{tabular*}{0.95\textwidth}{@{\extracolsep{\fill}} llcccccc }
        \toprule
        \textbf{Model} & \textbf{\#Data} & \textbf{MMMU$_{val}$} & \textbf{HallusionBench} & \textbf{MathVerse*} & \textbf{MMStar} & \textbf{MathVista}\\
        \midrule
        GPT-4o~\cite{hurst2024gpt}   & - & 60.0 & 68.0 & - & - & 63.8\\
        Gemini2-Flash~\cite{team2023gemini}&- & 70.6 & 69.4 & - & - & 70.4 \\
        Qwen2.5-VL-72B~\cite{Bai2025Qwen25VLTR}&- & 68.2 & 71.4 & - & 70.8 & 74.8 \\
        Qwen2.5-VL-32B~\cite{Bai2025Qwen25VLTR}&- & 63.7 & 72.1 & 54.3 & 67.3 & 74.7 \\
        \midrule
        InternVL2.5-8B~\cite{chen2024expanding}&- & 56.2 & 67.4 & - & 62.9  & 64.4\\
        MM-Eureka-7B~\cite{meng2025mm}  & 15K    & 55.4 & 69.5 & 46.6 & 64.6 & 73.0 \\
        OpenVLThinker-7B~\cite{deng2025openvlthinker} & 12K & 56.3 & 66.9 & 40.4 & 62.1 & 70.2 \\
        Perception-R1-7B~\cite{xiao2025perception} & 2K & 56.3 & 70.0 & 46.1 & 66.3 & 73.6\\
        Qwen2.5-VL-7B~\cite{Bai2025Qwen25VLTR} & -  & 56.4 & 63.8 & 42.6 & 64.3 & 68.2\\
        + GRPO & 4.7K & 57.8 & 66.7 & 45.9 & 66.0 & \textbf{74.1} \\
        PaLMR-7B & 4.7K  & \textbf{59.3} & \textbf{70.9} & \textbf{47.5} & \textbf{67.1} & 73.8 \\
        \bottomrule
    \end{tabular*}
    \caption{\textbf{Performance comparison of MLLMs on a suite of out-of-domain benchmarks.} Accuracy are reported for all benchmarks. MathVerse use vision only subset. Open-source models are evaluated using VLMEvalkit, and R1-style reasoning models are used  the same template as it training if provided.}
    \label{tab:main_results}
\end{table*}

\section{Experiments}
\subsection{Experimental Setup}\label{sec:exprm_setting}

We briefly introduce the training dataset, implementation details, baselines, and evaluation settings in this section.

\begin{figure}[htp]
    \centering
 \includegraphics[width=1.0\linewidth]{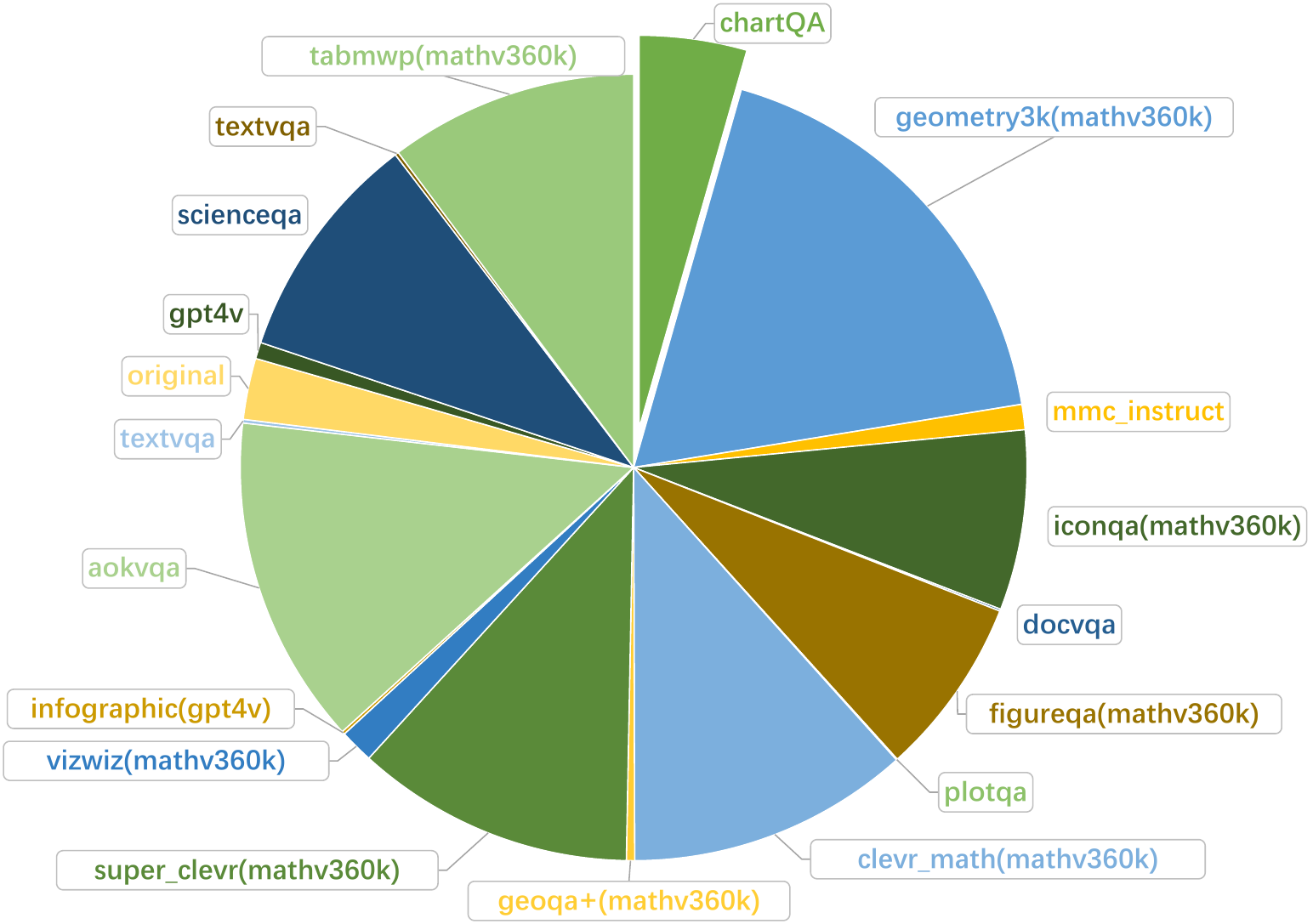}
    \caption{\textbf{Data distribution after our PaDLayer data filtering}. 19 distinct sub-domains are selected, and 4728 samples are finally used to generate the training dataset.}
    \label{fig:placeholder}
\end{figure}

\textbf{Training Dataset:}
As stated in Secion ~\ref{sec:PaDlayer}, we curated data from 19 distinct domains within the FineVision~\cite{wiedmann2025finevisionopendataneed} dataset, spanning areas such as geometry, Chart, Science, OCR, and general VQA. For each domain, we first sampled approximately 1.5K instances and discarded samples with low image-question relevance. The remaining candidates underwent a learnability-based data filtering step, resulting in a final training set of 4,728 high-quality instances. After that, we augment the standard VQA triplet(image, question, answer) with structured pseudo visual ground-truth and a reference response, providing richer guidance for process alignment training.

\textbf{Implementaion Details:}
We select a widely used multimodal LLM, Qwen2.5-VL-7B~\cite{Bai2025Qwen25VLTR}, as ~\textbf{PaLMR}'s backbone and trained it with the VeRL~\cite{seed2025verl}. In each reinforcement learning loop, we use Qwen3-30B-A3B~\cite{yang2025qwen3} as the percept-score judging LLM. All experiments were conducted on a cluster with 8 NVIDIA H100-80 GPUs. We use EasyR1 default parameters, but set the learning rate to 1e-6, the batch size to 128, the rollout batch size to 512, and the temperature to 1.0 during rolling. \textbf{V-GRPO} group size($G$) is set to 16, and the model is trained for 20 epochs with KL loss disabled. For reward calculation, we set the score coefficient $\alpha$ to 0.9 to emphasize the visual score of the final answer.

\textbf{Baseline approaches:}
To build a comprehensive evaluation of our method, we compare it with a vanilla GRPO training baseline and basemodel without training. Additionally, we include state-of-the-art MLLMs and similar open-source R1-style MLLMs: (1) Proprietary MLLMs: GPT4o~\cite{hurst2024gpt}, Gemini~\cite{team2023gemini}, (2) Open-source general MLLMs: Qwen2.5-VL series, InternVL 2.5 series~\cite{chen2024expanding}, (3) Open-source Reasoning MLLMs: MM-Eureka~\cite{meng2025mm}, OpenVLThinker~\cite{deng2025openvlthinker}, and Perception-R1~\cite{xiao2025perception}, All of which use Qwen2.5-VL-7B as the MLLM backbone, similar to our setting.

\textbf{Evaluation Benchmarks:} To demonstrate the enhanced reasoning performance of \textbf{PaLMR} across different domains, we employ a comprehensive set of visual math reasoning tasks and general multi-domain perception-centered reasoning tasks. These benchmarks include HallusionBench~\cite{Guan_2024_CVPR_hallusionBench} and MMStar~\cite{chen2024we}, which assess the model's robustness and performance on perception-intensive tasks that require fine-grained visual grounding on massive domains. Additionally, MMMU~\cite{yue2024mmmu}, MathVista~\cite{lu2023mathvista}, and the MathVerse~\cite{zhang2024mathverse} vision-only subset are focus problems demanding complex logical deductions from diagrams and images.

\subsection{Main Results}

Table~\ref{tab:main_results} shows that \textbf{PaLMR} achieves the best results on most out-of-distribution benchmarks among 7B-scale MLLMs. It not only surpasses general-purpose open-source MLLMs such as InternVL-8B across all benchmarks, but also outperforms reinforcement-learning-based reasoning models, including MM-Eureka and Perception-R1.


\textbf{The PaDLayer data curation pipeline demonstrates strong data efficiency.} Without visual-aware scores, the GRPO baseline (second-to-last row in Table~\ref{tab:main_results}) outperforms OpenVLThinker across all benchmarks, requiring only 4.7K training examples compared to 12K for OpenVLThinker, a 2.5-fold reduction under identical GRPO training conditions. When a process-level perception-aware score is incorporated into V-GRPO, the PaOLayer achieves state-of-the-art performance among 7B-scale models. PaLMR surpasses MM-Eureka-7B on MathVerse (47.5 vs 46.6) and HallusionBench (70.9 vs 69.5), even though vanilla GRPO does not outperform Eureka. These findings highlight the critical role of process-level visual alignment in GRPO: process-level visual scoring enforces consistency and reduces hallucinated reasoning (Figure~\ref{fig:teaser}).


\textbf{PaOLayer achieves strong scalability through noise robustness.} We compare against Perception-R1, which also integrates visual-aware scores into RL. Perception-R1 relies on point-wise scoring and requires Gemini to generate complex, question-relevant visual content. In contrast, PaOLayer employs pairwise reranking and only requires Gemini to produce a generic image caption. Despite relying on these less specialized data, PaLMR outperforms Perception-R1 on all benchmarks. This confirms that pairwise scoring, by leveraging self-generated reference responses, reduces reliance on high-quality, task-specific annotations and offers better scalability potential.

\begin{figure}[htp]
    \centering
    \includegraphics[width=1.0\linewidth]{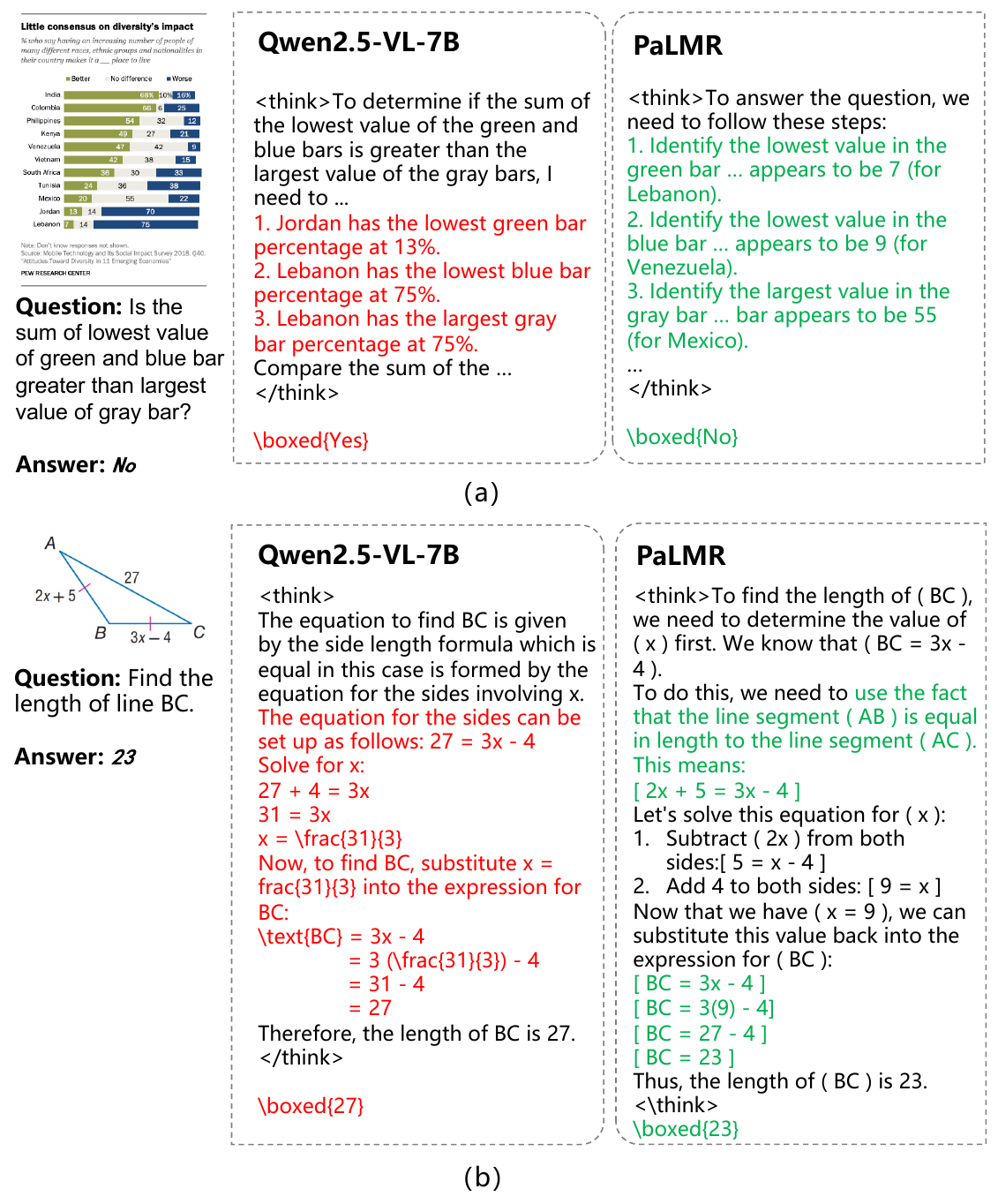}
    \caption{\textbf{Qualitative comparison of reasoning chains between baseline models and PaLMR across different domains.}}
    \label{fig:vis_result_1}
\end{figure}

Qualitative comparisons (Figure~\ref{fig:vis_result_1}) further validate that PaLMR consistently produces visually aligned reasoning chains while maintaining answer correctness. Ultimately, process-level perception alignment improves not only benchmark performance but also reasoning stability, as further analyzed in Figure~\ref{fig:response_length}.

\subsection{Generalizability Across Model Scales and Architectures}

To validate the robustness and scalability of PaLMR, we extend our evaluation beyond the base 7B model to different parameter scales and newer architectural generations. The comprehensive results are summarized in Table~\ref{tab:size_study}.

Across the evaluated model scales ranging from 3B to 32B on the Qwen2.5 family, PaLMR consistently outperforms the GRPO method on out-of-domain benchmarks. In particular, on the 32B model, PaLMR improves the baseline MMMU performance from 64.3 to 66.8, while GRPO slightly degrades to 64.0. This confirms that PaLMR generalizes well to models within the same family. 

While PaLMR yields substantial improvements on the Qwen2.5-VL series, its performance-boosting effect diminishes when applied to the more advanced Qwen3-VL-8B architecture. Specifically, on the HallusionBench and MathVerse Vision Only benchmarks, PaLMR achieves scores of 75.2 and 60.0, marginally underperforming the standard GRPO baselines of 75.3 and 60.8. This performance saturation can be attributed to the degradation of the visual reward mechanism. PaLMR utilizes a pair-wise scoring paradigm based on reference data annotated by the less capable Qwen2.5-VL-7B model. As the target model's intrinsic capabilities eclipse those of the annotator, the judge model inherently loses its discriminative precision. Consequently, the visual gate mechanism fails to accurately quantify performance gains, causing the optimization framework to degenerate into standard GRPO.

\begin{table}[tp]
    \centering
    \vspace{-10pt}
    \resizebox{\linewidth}{!}{
        \begin{tabular}{lccccc}
            \toprule
            \textbf{Model} & \textbf{MMMU} & \textbf{Hallusion} & \textbf{MathVerse} & \textbf{MMStar} & \textbf{MathVista}\\
            \midrule
            Qwen2.5-VL-3B & 49.3 & - & 32.4 & 55.0 & 63.3 \\
            + GRPO & 53.3 & - & 34.0 & 56.7 & 62.8 \\
            PaLMR-3B & \textbf{53.9} & - & \textbf{38.8} & \textbf{57.2} & \textbf{64.0} \\
            \midrule
            Qwen2.5-VL-7B &  56.4 & 63.8 & 42.6 & 64.3 & 68.2\\
            + GRPO &  57.8 & 66.7 & 45.9 & 66.0 & \textbf{74.1} \\
            PaLMR-7B &  \textbf{59.3} & \textbf{70.9} & \textbf{47.5} & \textbf{67.1} & 73.8 \\
            \midrule 
            Qwen2.5-VL-32B$^\dagger$ & 64.3 & 69.1 & 49.4 & 65.8 & 75.1 \\ 
            + GRPO$^\dagger$ & 64.0 & 70.1 & 49.7 & 66.9 & 72.9 \\
            PaLMR-32B$^\dagger$ & \textbf{66.8} & \textbf{71.5} & \textbf{51.3} & \textbf{67.6} & 74.5 \\
            \midrule
            Qwen3-VL-8B$^\dagger$ & 61.3 & 73.5 & - & 70.9 & 77.2 \\
            + GRPO$^\dagger$ & 64.6 & \textbf{75.3} & \textbf{60.8} & 72.6 & 77.5 \\
            PaLMR-8B$^\dagger$ & \textbf{65.3} & 75.2 & 60.0 & 72.6 & \textbf{78.3} \\
             
            \bottomrule
        \end{tabular}
    }
    \caption{\textbf{Comparison of model performance across different scales on out-of-domain benchmarks. } 
$^\dagger$ indicates inference via vLLM.}
    \label{tab:size_study}
    \vspace{-10pt}
\end{table}

\subsection{Effectiveness of the Perception-Aware Score}\label{sec: perception_diff}

\begin{figure}[tp]
    \centering
    \includegraphics[width=0.98\linewidth]{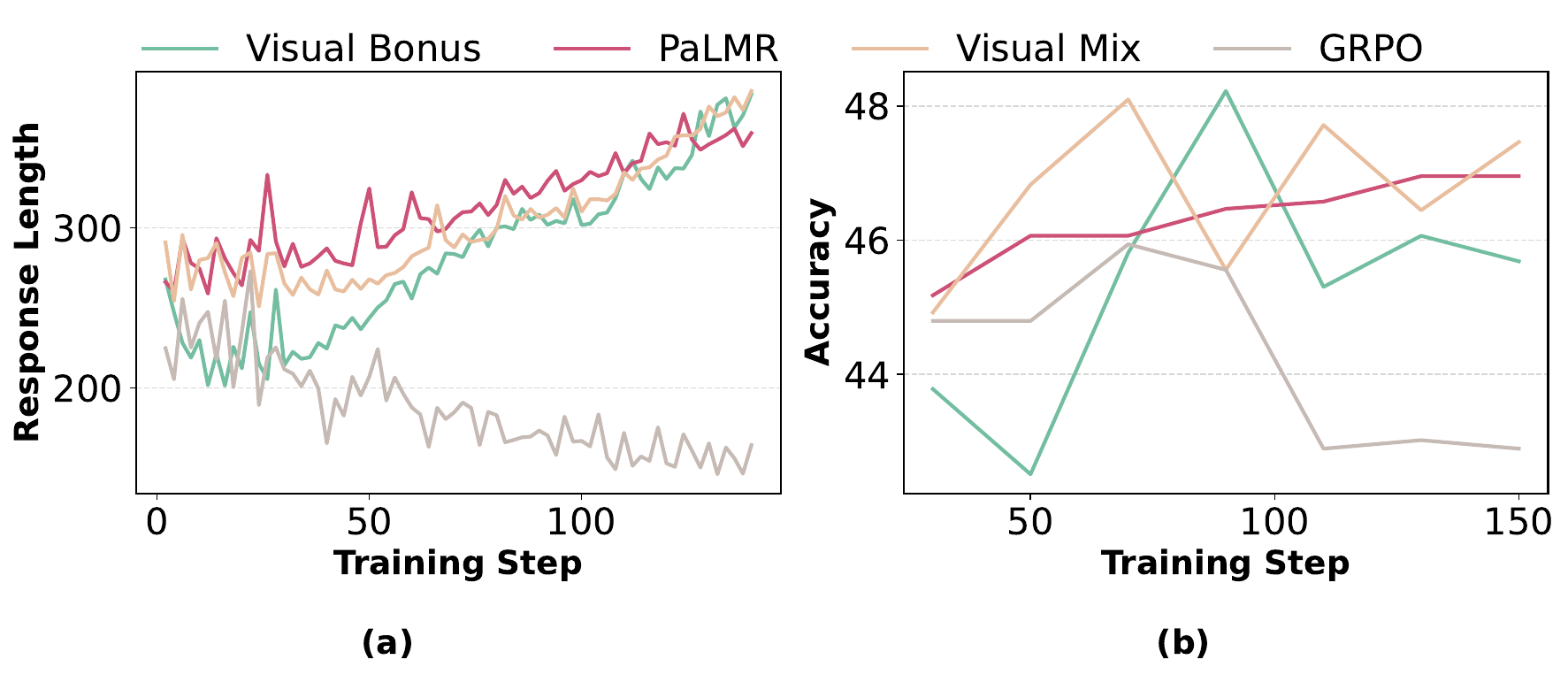}
    \caption{\textbf{(a) Comparison of average response length on the training set across different reward settings. (b) Comparison of accuracy on the MathVerse vision-only benchmark.}}
    \label{fig:response_length}
\end{figure}

\begin{table}[tp]
    \centering
    \resizebox{\columnwidth}{!}{
    \begin{tabular}{lcccc}
        \toprule
        \textbf{Model} & \textbf{MMMU$_{val}$} & \textbf{MathVerse*} & \textbf{MMStar} & \textbf{MathVista}\\
        \midrule
        Qwen2.5-VL-7B  & 56.4 & 42.6 & 64.3 & 68.2\\
        + GRPO & 57.8  & 45.9 & 66.0 & \textbf{74.1} \\
        Visual Bonus-7B  & 59.3  & \textbf{48.2} & 65.0 & 73.9 \\
        Visual Mix-7B  & 58.6  & 48.1 & 66.5 & 72.5 \\
        \textbf{PaLMR}-7B & \textbf{59.3}  & 47.5 & \textbf{67.1} & 73.8 \\
        \bottomrule
    \end{tabular}
    }
    \caption{\textbf{An ablation study on the formulation strategies of visual-aware rewards.} Utilizing the identical configuration for training as the primary experiment, we compare the proposed \textbf{PaLMR} against the strategies of Visual Bonus and Visual Mix across standard benchmarks.}
    \label{tab:visual_signal_combination}
\end{table}

To systematically evaluate the integration of visual perception signals with the correctness of the final answer, we unify and compare multiple strategies for reward formulation. These strategies combine the visual alignment score $S_{p,vis}$, the answer correctness score $S_{p,ans}$, and the formatting score $S_{p,fmt}$ in distinct configurations. Specifically, we design three representative variants within a standard GRPO training framework:

\begin{enumerate}
    \item \emph{Vanilla GRPO} (baseline) and \textbf{PaLMR} adopt the settings detailed in Sec.~\ref{sec:PaOLayer}.
    \item \emph{Visual Bonus}. This strategy introduces a supplementary reward bonus when the model achieves both correct answers and visual alignment with the ground truth. The reward is formulated as $R = \alpha S_{p,ans} + (1-\alpha) S_{p,fmt} + C$, where $C = 0.5$.
    \item \emph{Visual Mix}. This strategy integrates the visual score as a weighted component of the overall reward, formulated as $R = \alpha S_{p,vis} + \beta S_{p,ans} + \gamma S_{p,fmt}$, where $\alpha + \beta + \gamma = 1$. For our experiments, we set $\alpha = 0.2$, $\beta = 0.7$, and $\gamma = 0.1$.
\end{enumerate}

This unified design facilitates a rigorous evaluation of different paradigms for integrating visual consistency with answer correctness within a single training pipeline.
Figure~\ref{fig:response_length} illustrates the training dynamics of the four aforementioned strategies, evaluated by the average length of responses and the accuracy on the vision-only subset of MathVerse.

\textbf{Impact on Response Generation:} 
As shown in Figure~\ref{fig:response_length}(a), the three strategies that incorporate the visual-aware score, namely \textbf{PaLMR}, Visual Mix, and Visual Bonus, consistently drive the model to generate longer responses than the baseline of vanilla GRPO. 
This observation indicates that providing rewards for visual alignment encourages the model to initiate the reasoning process with visual analysis, even in the absence of explicit prompts. 
Conversely, the baseline of GRPO tends to generate shorter responses that directly target the final answer.

\textbf{Performance and Stability Trade-off:} As shown in Figure~\ref{fig:response_length}(b), PaLMR maintains a stable, monotonically increasing accuracy on MathVerse. This stability stems from its hierarchical reward, which serves as a rigorous gating mechanism that requires coherent visual perception before rewarding the final answer ($S_{p,ans}$). In contrast, treating the visual score as a mere auxiliary objective (Visual Mix/Bonus) leads to volatile training, as models bypass visual grounding in favor of spurious, result-oriented solutions.

Table~\ref{tab:visual_signal_combination} highlights this critical trade-off: for base models with limited capacity, visual faithfulness and outcome correctness often diverge, especially on reasoning-heavy tasks. While PaLMR's strict visual gating enforces alignment to ensure superior robustness on comprehensive benchmarks (MMMU, MMStar), auxiliary objectives (Visual Mix/Bonus) exploit this divergence. They achieve marginally higher peak accuracy on specific math tasks (MathVerse) by bypassing visual verification.

\subsection{Error Analysis and Reward Robustness}
\label{sec:error_analysis}

\begin{figure}[tp]
    \centering
    \includegraphics[width=0.98\linewidth]{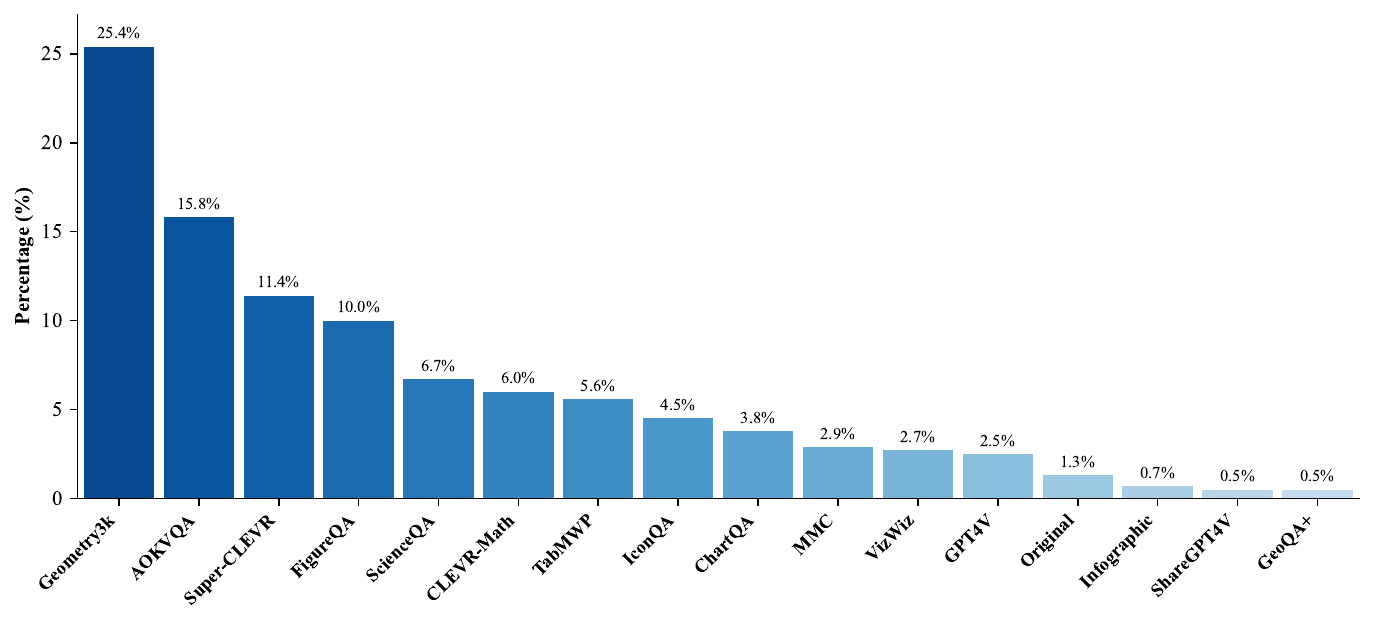}
    \caption{\textbf{Distribution of persistent errors across data sources on the training set.}}
    \label{fig:train_err_distribution}
\end{figure}

To understand the boundaries of PaLMR, we analyze its persistent training failures. As shown in Figure~\ref{fig:train_err_distribution}, over half of these errors originate from three datasets: Geometry3k~\cite{lu2021inter-geo3k} (25.4\%), AOKVQA~\cite{schwenk2022okvqa} (15.8\%), and Super-CLEVR~\cite{li2023super-clevr} (11.4\%). Manual inspection categorizes these failures into two primary mechanisms:

First, \textbf{modality limitations} (e.g., Geometry3k): pure textual reasoning inherently struggles to capture complex, implicit spatial and geometric constraints. Second, \textbf{overly strict evaluation} (e.g., AOKVQA): exact-matching introduces false penalties for semantically valid but lexically distinct descriptions (e.g., "blue" vs. "cyan").

However, this strict exact-matching acts as a double-edged sword, serving as a \textbf{robust filter against label noise}. Synthetic datasets like Super-CLEVR and CLEVR-Math~\cite{lindstrom2022clevr-math} frequently contain visually inconsistent ground truths. While the standard GRPO baseline hallucinates incorrect reasoning paths to overfit these flawed labels, PaLMR's intrinsic consistency gating resists this deceptive optimization pressure. By actively rejecting corrupted synthetic samples, PaLMR effectively preserves the mathematical and logical soundness of its outputs.




\subsection{Limitations}
While PaLMR establishes a robust foundation for process-aligned multimodal reasoning, we identify two primary directions for ongoing exploration. First, PaLMR's reasoning upper bound is constrained by the base model's foundational capabilities; future work could explicitly integrate auxiliary visual experts to improve this perceptual baseline. Second, our exact-matching reward inadvertently penalizes lexically distinct but semantically valid descriptions. Formulating robust, semantic-aware consistency metrics is necessary to reduce these false penalties.

\section{Conclusion}
We present PaLMR (Process Alignment for Multimodal Reasoning), a unified framework for faithful multimodal process-level reasoning. In contrast to previous reinforcement learning approaches that focus solely on final-answer correctness, PaLMR enforces alignment across perception, reasoning, and optimization, ensuring visually consistent and interpretable reasoning. The perception-aligned data construction pipeline, PaDLayer, programmatically generates multimodal datasets with structured and verifiable textual descriptions, providing a reliable basis for evaluating process-level faithfulness. Building on this, we introduce a process-aligned optimization layer, PaOLayer, instantiated with Vision-Guided GRPO (V-GRPO), which incorporates hierarchical perception-aware scores into the reward to promote visual consistency and reasoning coherence during reinforcement learning. Empirical results across multiple multimodal reasoning benchmarks show that PaLMR, trained with approximately 4.7K high-quality multimodal samples, consistently improves visual faithfulness and reasoning stability while maintaining competitive accuracy. These findings demonstrate that aligning the reasoning process itself, rather than optimizing only for outcome correctness, is essential for developing more reliable and interpretable multimodal large language models.

\clearpage

\section*{Acknowledgement}

This work was supported by the National Natural Science Foundation of China Enterprise Innovation and Development Joint Fund Project U24B20181. The authors extend their sincere gratitude to Wenjie Qiu and Wenpo Song for their constructive discussions and continuous support throughout this project.

{
    \small
    \bibliographystyle{ieeenat_fullname}
    \bibliography{main}
}
\clearpage

\appendix
\noindent {\Large \textbf{Appendix}} 


\section{Training Data Distribution}


Table \ref{tab:dataset_detailed} delineates the domain-specific distribution of the filtered training dataset.
Our filtering protocol was designed to remaining samples learnable for the reinforcement learning stage.
First, because we employ a rule-based direct matching method to evaluate response correctness, tasks that inherently require open-ended or verbose optical character recognition (OCR) outputs were removed.
Consequently, the representation of OCR-heavy datasets (e.g., DocVQA) is intentionally minimized. 
Furthermore, we filtered samples based on model solvability to prevent hallucinations on tasks exceeding the capacity of our base model, Qwen2.5-VL-7B. 
Specifically, we excluded instances from PlotQA (complex plot reasoning), VizWiz (fine-grained grounding for visually impaired users), and human-annotated MMC-Instruct that required advanced real-world chart understanding, etc.
This ensures that the training data remains within the learnable manifold of the base model while maintaining diversity across geometry, science, and chart domains.

\begin{table}[h]
\centering
\resizebox{\columnwidth}{!}{
\begin{tabular}{llrr}
\toprule
Category & Dataset & Number of Samples & Percentage (\%) \\
\midrule
\textbf{Chart VQA} & ChartQA\cite{masry2022chartqa} & 209 & 4.42 \\
 & FigureQA\cite{kahou2017figureqa} & 348 & 7.36 \\
 & PlotQA\cite{methani2020plotqa} & 2 & 0.04 \\
 & TabMWP\cite{lu2022dynamictabmwp} & 483 & 10.22 \\
 & \textbf{Subtotal} & \textbf{1042} & \textbf{22.04} \\
\cmidrule{1-4}
\textbf{GeoVQA} & GeoQA+\cite{cao-xiao-2022-augmented-geoqa+} & 16 & 0.34 \\
 & Geometry3K\cite{lu2021inter-geo3k} & 852 & 18.02 \\
 & \textbf{Subtotal} & \textbf{868} & \textbf{18.36} \\
\cmidrule{1-4}
\textbf{Science VQA} & ScienceQA\cite{lu2022learn-sciqa} & 449 & 9.50 \\
 & \textbf{Subtotal} & \textbf{449} & \textbf{9.50} \\
\cmidrule{1-4}
\textbf{Math VQA} & CLEVR-Math\cite{lindstrom2022clevr-math} & 549 & 11.61 \\
 & Super-CLEVR\cite{li2023super-clevr} & 542 & 11.46 \\
 & IconQA\cite{lu2021iconqa} & 350 & 7.40 \\
 & \textbf{Subtotal} & \textbf{1441} & \textbf{30.48} \\
\cmidrule{1-4}
\textbf{OCR VQA} & DocVQA\cite{mathew2021docvqa} & 4 & 0.08 \\
 & TextVQA\cite{singh2019towards-textvqa} & 7 & 0.15 \\
 & InfographicVQA\cite{mathew2022infographicvqa} & 6 & 0.13 \\
 & \textbf{Subtotal} & \textbf{17} & \textbf{0.36} \\
\cmidrule{1-4}
\textbf{General VQA} & A-OKVQA\cite{schwenk2022okvqa} & 640 & 13.54 \\
 & VizWiz\cite{bigham2010vizwiz} & 66 & 1.40 \\
 & MMC-Instruct\cite{liu-etal-2024-mmc} & 49 & 1.04 \\
 & Original & 119 & 2.52 \\
 & GPT-4V & 32 & 0.68 \\
 & ShareGPT4V\cite{chen2024sharegpt4v} & 5 & 0.11 \\
 & \textbf{Subtotal} & \textbf{911} & \textbf{19.27} \\
\cmidrule{1-4}
\textbf{Total} & & \textbf{4728} & \\
\bottomrule
\end{tabular}
}
\caption{Detailed Dataset Composition. The distribution represents the training data after applying solvability and evaluability filtering heuristics. The `Subtotal' rows indicate the aggregate number of each domain.}
\label{tab:dataset_detailed}
\end{table}

\section{Prompt Templates for PaLMR}

\begin{figure}[hp]
    \centering
    \includegraphics[width=0.95\linewidth]{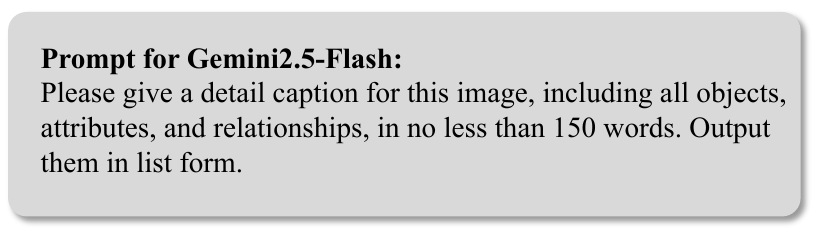}
    \caption{Prompt template for Visual Ground Truth Generation. The prompt template used to instruct Gemini-2.5-Flash to generate structured, question-agnostic visual descriptions.}
    \label{fig:caption_prompt}
\end{figure}

\textbf{Pseudo-Visual Ground Truth Generation.} 
Obtaining large-scale, human-annotated visual descriptions as ground truth is too expensive. To address this, we leverage Gemini-2.5-Flash for its optimal trade-off between inference speed and visual perception capability.
As illustrated in Figure \ref{fig:caption_prompt}, Gemini is prompted to generate structured descriptions of the visual content directly.
Crucially, this generation is performed in a question-agnostic manner. By decoupling the description from the specific query, we ensure that the visual ground truth captures a comprehensive representation of the image states. To ensure scalability and computational efficiency, particularly for datasets where a single image is associated with multiple question-answer pairs (one-to-many mapping)

\begin{figure}[htp]
    \centering
    \includegraphics[width=0.95\linewidth]{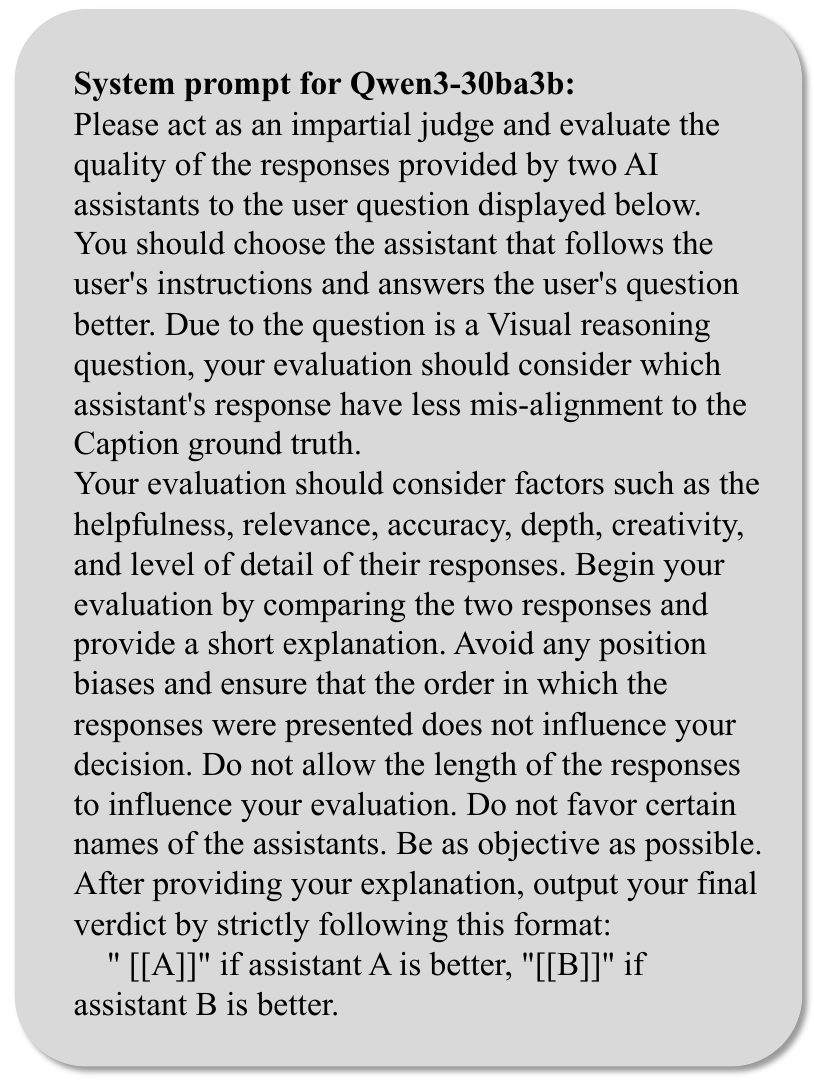}
    \caption{System prompt template for Visual-Aware Scoring. The LLM-as-a-Judge prompt used by Qwen3-30ba3b to perform pairwise ranking between the model rollout and a reference response, conditioned on the pseudo-visual ground truth.}
    \label{fig:judge_prompts_sys}
\end{figure}

\begin{figure}[ht]
    \centering
    \includegraphics[width=0.95\linewidth]{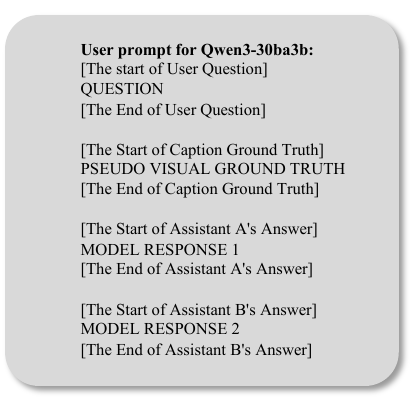}
    \caption{User prompt template for Visual-Aware Scoring.}
    \label{fig:judge_prompts_user}
\end{figure}

\begin{figure}[h]
    \centering
    \includegraphics[width=0.95\linewidth]{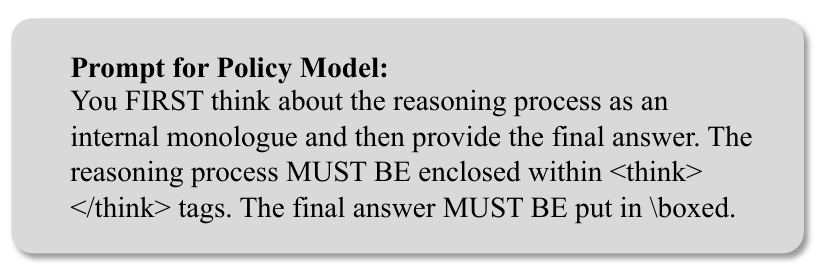}
    \caption{Prompt template for Policy Model.}
    \label{fig:think_prompt}
\end{figure}

\textbf{Visual-Aware Scoring Mechanism.} 
To verify visual fidelity during the reasoning process, we implement an LLM-as-a-Judge metric on the training set.
We employ Qwen3-30ba3b as the judge model to balance evaluation accuracy with computational efficiency.
To align the reward signal with human judgment, we formulate the evaluation as a pairwise re-ranking task.
As shown in Figure \ref{fig:judge_prompts_sys} and \ref{fig:judge_prompts_user}, the judge is provided with the question, the generated pseudo-visual ground truth (from Gemini), the current model rollout, and a pre-selected reference response from the base model.
The judge evaluates the current rollout against the reference, explicitly conditioning its decision on the structured visual ground truth. This pipeline significantly improves the consistency of the scoring and the alignment rate with human preference.

\textbf{Thinking prompt.}
We design our prompt template following the format in EasyR1, wherein the user
prompt explicitly specifies the required output structure, including the use of \verb|<think><\think>| and \verb|\boxed{}| tags to separate the reasoning process and the final answer as in Figure \ref{fig:think_prompt}.
This prompt is appended to all queries for training samples, not set as a system prompt.

\section{Implementation Details for Visual-Aware Scoring}

\textbf{Inference Optimization and Verdict Extraction.}
We observe that the thinking token used for PaLMR (e.g., \verb|<think>|) will trigger Qwen3's thinking mode.
While useful for complex reasoning, this introduces significant latency, rendering it computationally prohibitive for online reinforcement learning.
To solve this problem, we remove all \verb|<think>| and \verb|<\think>| tokens during the judge's generation phase.

Furthermore, despite prompt instructions requesting a structured format, large language models occasionally produce free-form text containing multiple potential verdict tokens. 
To address this, we adopt a robust extraction method similar to those employed in multimodal benchmarks like MMMU. Specifically, we utilize regular expressions to identify all valid verdict candidates within the response. We select the \textit{last} matching token as the final decision, exploiting the tendency of instruction-tuned models to output their definitive conclusion at the end of the reasoning trace.

\textbf{Mitigating Positional Bias via Input Shuffling.}
LLM-as-a-Judge metrics are susceptible to positional bias, where the model exhibits a preference for options presented in specific locations (e.g., "Option A" vs. "Option B") regardless of content quality.
To ensure accuracy, we implement a stochastic rolling strategy for input construction. For each evaluation pair consisting of the model's current rollout and the reference response, we randomize their assignment to the first or second position in the prompt context. The judge's output is then dynamically mapped back to the corresponding source (model or reference) to derive the final visual fidelity score.

\section{Visualized Examples for Training Dataset}

\begin{figure*}
    \centering
    \includegraphics[width=0.95\linewidth]{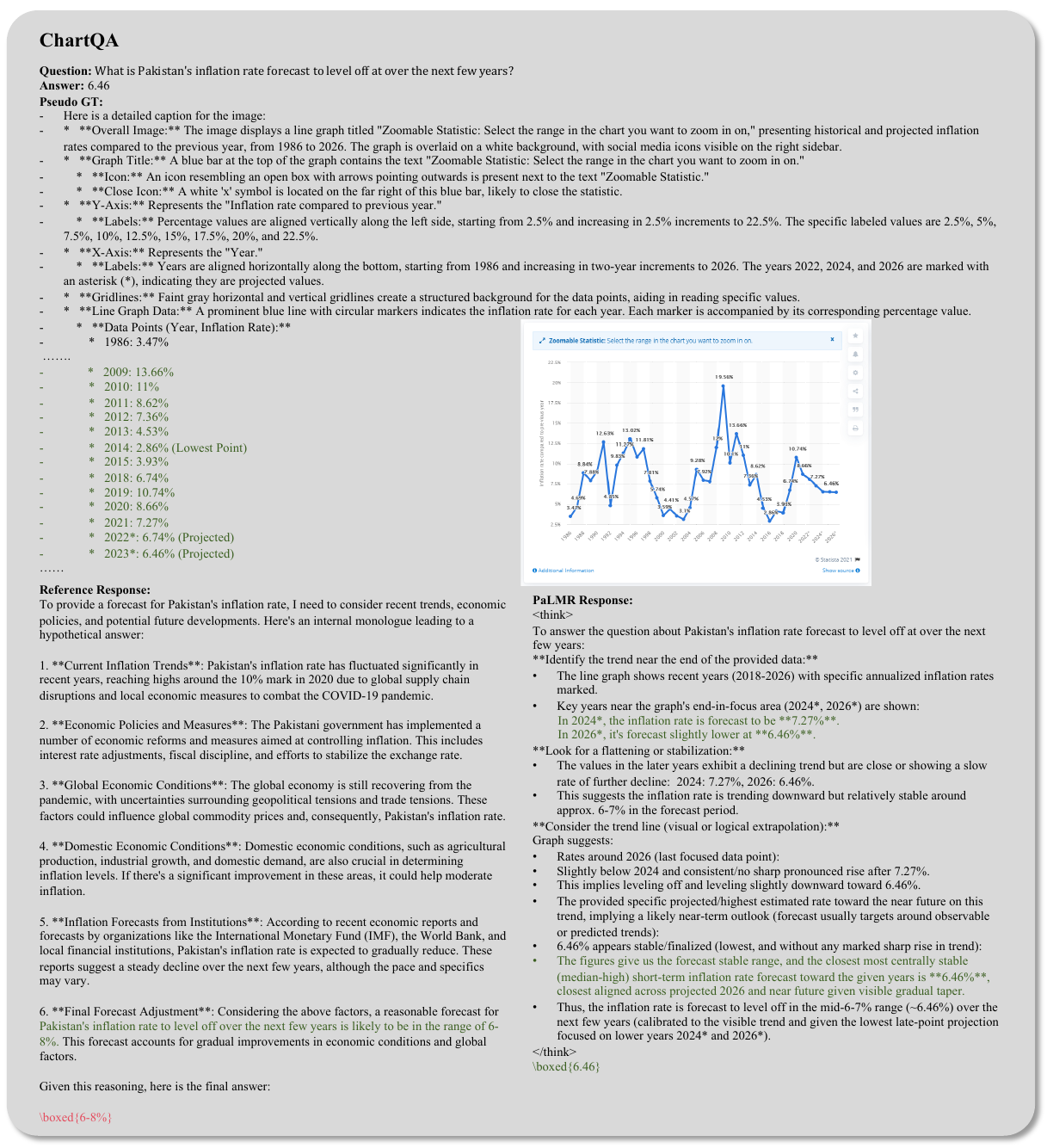}
    \caption{Visualization of training set.}
    \label{fig:train_dd1}
\end{figure*}

\begin{figure*}
    \centering
    \includegraphics[width=0.95\linewidth]{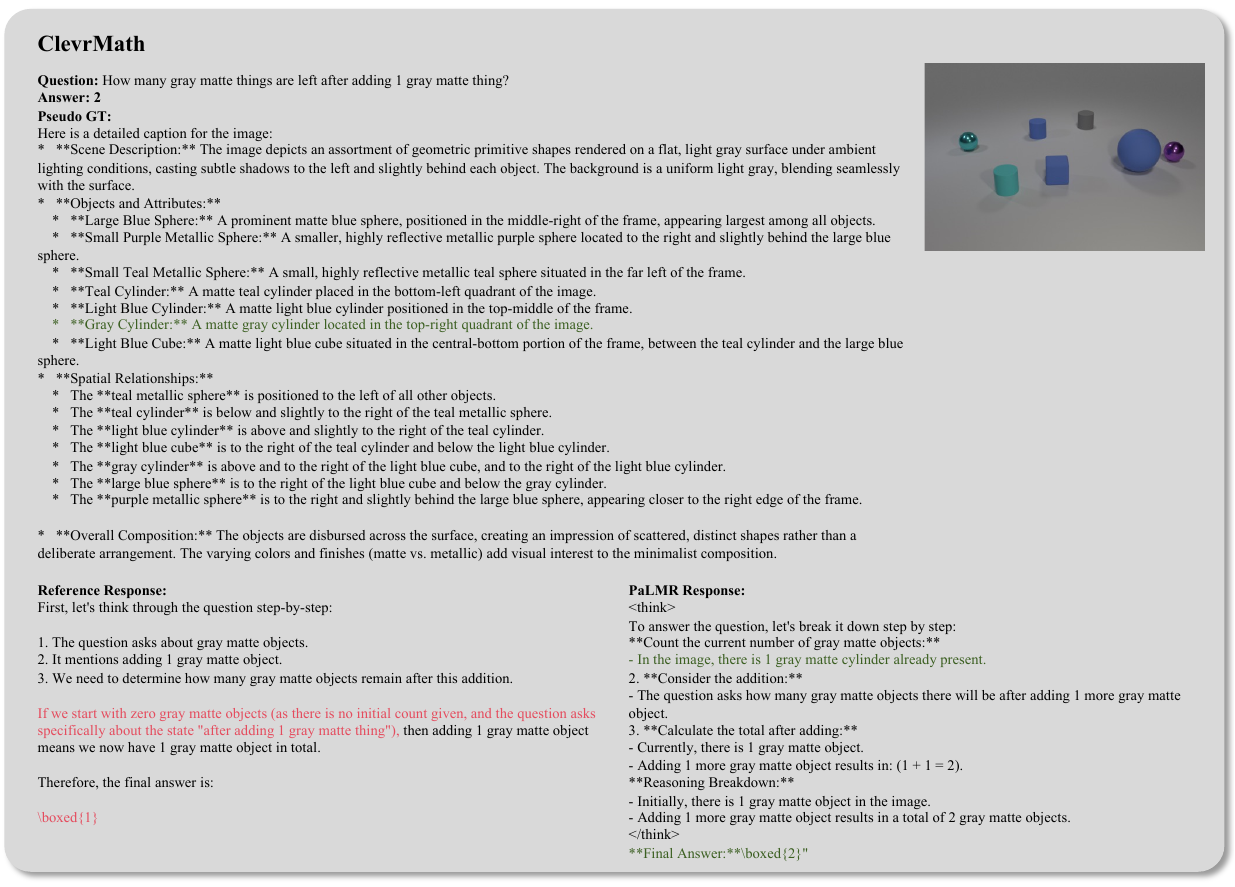}
    \caption{Visualization of training set.}
    \label{fig:train_dd2}
\end{figure*}

Figures \ref{fig:train_dd1} and \ref{fig:train_dd2} present qualitative examples from our training dataset alongside the inference outputs of the trained PaLMR model. In these visualizations, red text denotes incorrect reasoning or hallucinations, whereas green text highlights visually accurate observations relevant to the user's query.

First, we observe that the pseudo-visual ground truth synthesized by Gemini-2.5-Flash consistently provides precise descriptions of visual content, capturing details essential for the reasoning process. This reliability validates the use of pseudo visual GTs within our visual-aware scoring framework as a robust metric for assessing visual faithfulness. Furthermore, while the base model suffers from visual inconsistencies—such as inaccurate data extraction in ChartQA and object recognition failures in ClevrMath—PaLMR effectively rectifies these hallucinations through visual guidance.
It ultimately yields responses that are both visually perceptive and logically correct.

\section{Computational Cost and Latency Analysis}
\label{sec:appendix_latency}

We analyze the computational overhead of our proposed PaLMR framework compared to the standard GRPO baseline. Table \ref{tab:latency} details the per-step training latency for Qwen2.5-VL-7B on a node with 8 GPUs, which are allocated symmetrically (4 GPUs for the policy model and 4 for the reward model).

By serving the judge model (Qwen3-30A3B) via SGLang, PaLMR introduces a 37.5\% latency overhead relative to the baseline. To investigate this cost, we conducted an ablation study by disabling the Chain-of-Thought (CoT) generation in the judge model (PaLMR w/o CoT). While this nearly eliminates the computational overhead (+0.7\%), it substantially degrades downstream performance, reducing the MMMU score from 59.3 to 57.3 and dropping human alignment to approximately 60\%. These findings demonstrate that the additional computational cost incurred by CoT is a necessary trade-off. The CoT process is essential for providing stable, low-noise reward signals that effectively mitigate judge model bias, thereby enabling superior reasoning capabilities.

\begin{table}[h]
    \centering
    \small
    \setlength{\tabcolsep}{3pt} 
    \vspace{-5pt} 
    \begin{tabular}{l c c c}
        \toprule
        \textbf{Method}  & \textbf{Latency (s)} & \textbf{Overhead} & \textbf{MMMU} \\
        \midrule
        Baseline (GRPO)  & 397 & +0\% & 57.8 \\
        PaLMR (w/o CoT)  & $$400 & $$ +0.7\% & 57.3 \\
        \textbf{PaLMR (Ours)}  & \textbf{546} & \textbf{+37.5\%} & \textbf{59.3}\\
        \bottomrule
    \end{tabular}
    \vspace{-10pt} 
    \caption{Comparison of per-step training latency, computational overhead, and downstream performance (MMMU). Evaluated on Qwen2.5-VL-7B using 8 GPUs.}
    \label{tab:latency}
\end{table}

\end{document}